\documentclass[10pt,twocolumn,letterpaper]{article}

\usepackage[pagenumbers]{cvpr} 

\usepackage{currfile} 

\usepackage{caption} 

\usepackage{amsthm}
\newtheorem{assumption}{Assumption}[section]

\usepackage{amsthm}
\newtheorem{theorem}{Theorem}

\usepackage{multirow}
\usepackage{tabularray}

\usepackage{algorithm}
\usepackage{algpseudocode}

\usepackage{colortbl}

\algrenewcommand\algorithmicrequire{\textbf{Input:}}
\algrenewcommand\algorithmicensure{\textbf{Output:}}

\definecolor{cvprblue}{rgb}{0.21,0.49,0.74}
\usepackage[pagebackref,breaklinks,colorlinks,allcolors=cvprblue]{hyperref}

\def\confName{CVPR}
\def\confYear{2027}

\title{CRISP: Skeleton-Aware Conformal Regions for Reliable \\Human Motion Prediction}

\author{Dong Wei\textsuperscript{1}, Liwen Chen\textsuperscript{1}, Xiaoning Sun\textsuperscript{2}, Yadang Chen\textsuperscript{3}, Fan Liu\textsuperscript{1}, Yuhui Zheng\textsuperscript{4}\\[2pt]
{\small\textsuperscript{1}Hohai University \enskip \textsuperscript{2}Nanjing Normal University \enskip \textsuperscript{3}Nanjing University of Information Science \& Technology \enskip \textsuperscript{4}Qinghai Normal University}}

\begin{document}
\frenchspacing
\maketitle
\begin{abstract}
Human motion prediction supports risk-aware decision-making in safety-critical applications such as human-robot collaboration. However, existing methods primarily emphasize predictive accuracy or motion diversity, neither of which can provide statistically guaranteed uncertainty bounds for assessing prediction reliability and mitigating potential risks. In this paper, we propose CRISP, a model-agnostic conformal prediction framework for uncertainty quantification in 3D human motion forecasting. This formulation transforms a pretrained motion predictor into a calibrated set-valued predictor, yielding compact prediction regions for the entire future motion without retraining. Under the stated assumptions, the resulting regions provide a finite-sample guarantee of simultaneous coverage across all body joints and future time steps. Spatially, we integrate skeletal topology priors with empirical prediction errors to construct geometry-aware uncertainty regions that capture inter-joint dependencies and reduce conservatism. Temporally, we employ copula-based calibration to exploit dependencies across forecasting horizons and jointly calibrate prediction regions without specifying a parametric error distribution. Experiments on four datasets demonstrate that CRISP achieves empirical joint coverage close to the target level while producing more compact prediction regions than representative conformal prediction baselines.
\end{abstract}    
\section{Introduction}
\label{sec:intro}
3D human motion prediction is a fundamental and challenging task in computer vision, aiming to forecast future 3D poses of the human body based on observed historical motion sequences. As a core component of understanding human behaviors, it has attracted significant attention in practical applications, including human-robot interaction \cite{koppula2015anticipating,vianello2021human,liu2026adaptive}, autonomous driving \cite{fernando2020deep}, and healthcare \cite{capogrosso2018configuration}.

Driven by the rapid advancements in deep learning, previous methods for human motion prediction mainly fall into two paradigms. The first paradigm focuses on deterministic motion prediction, striving to maximize point-estimation accuracy. By leveraging dedicated network architectures (e.g., RNNs \cite{Res-sup}, GCNs \cite{ltd,li2022skeleton}, Transformers \cite{aksan2021spatio}, MLPs \cite{guo2023back}), these methods deeply mine the complex spatial-temporal structures inherent in human movements, yielding a single, highly precise future motion sequence. 
Recognizing that human motion is intrinsically multimodal and uncertain, the second paradigm shifts towards stochastic motion prediction. These approaches incorporate generative modeling techniques (e.g., VAEs \cite{yuan2020dlow}, GANs \cite{zhao2023bidirectional}, DDPMs \cite{chen2023humanmac}) to generate a diverse set of plausible future hypotheses, aiming to capture the inherent diversity of human behaviors while maintaining prediction accuracy.

\begin{figure}[t]
\centering
\includegraphics[width=1.0\columnwidth]{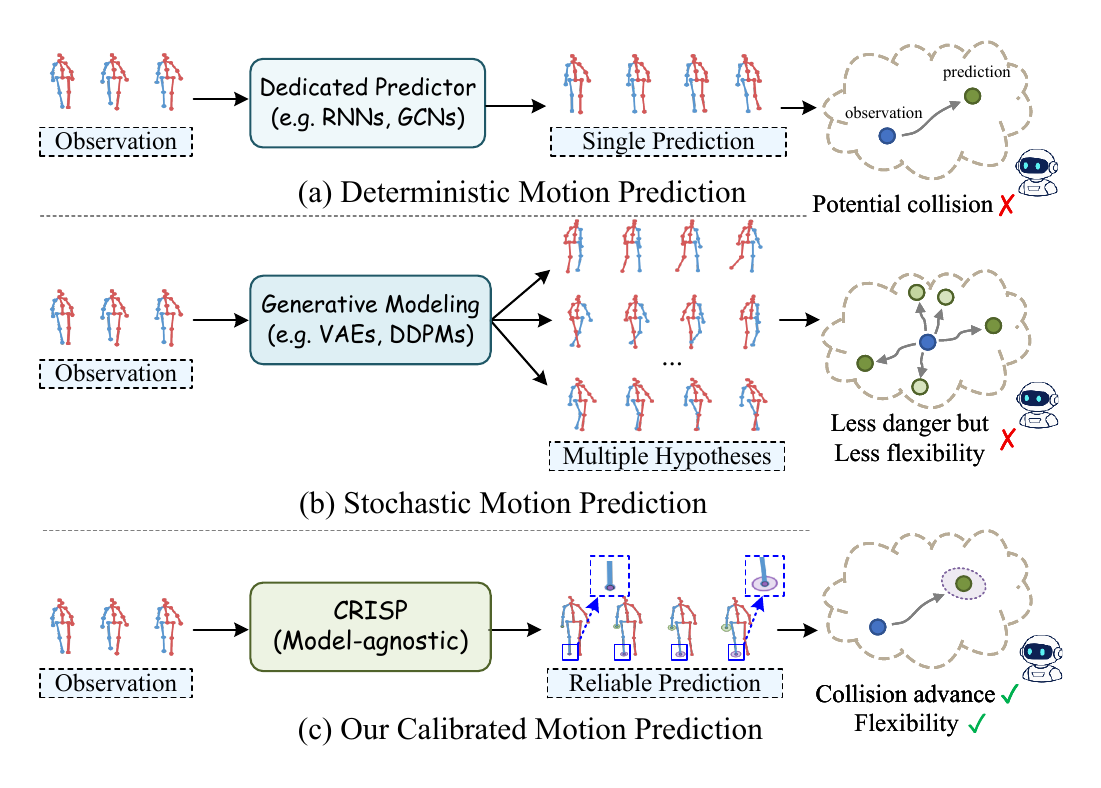}
\vspace{-9mm}
\caption{Comparison of motion prediction paradigms. Unlike (a) deterministic methods that completely ignores uncertainty, and (b) stochastic methods predicting multiple hypotheses with unbounded variance, (c) our CRISP produces reliable predictions with mathematically guaranteed uncertainty bounds, ensuring safe and trustworthy deployment.
From the robot's perspective, these reliable bounds facilitate potential collision risk assessment while preserving motion flexibility.}
\vspace{-4mm}
\label{fig:1}
\end{figure}

However, deploying these paradigms in real-world, safety-critical applications poses significant risks due to the absence of reliable and calibrated uncertainty bounds. For instance, in human-robot interaction, a collaborative robot must anticipate the strict boundaries of a human's potential movements to plan risk-controlled trajectories; relying on a single deterministic guess or an unconstrained set of stochastic hypotheses could lead to catastrophic collisions. While recent works \cite{wang2024existence,saadatnejad2024toward} have taken initial steps to incorporate parametric uncertainty modeling into pose forecasting, they are fundamentally bottlenecked by their reliance on heuristic variance estimations or rigid parametric assumptions (e.g., Gaussian distributions). Since real-world human dynamics are inherently complex and strictly non-Gaussian, forcing these dynamics into predefined statistical templates inevitably yields miscalibrated prediction regions. More critically, they lack rigorous coverage guarantees, rendering their estimated boundaries mathematically ungrounded and unreliable for trustworthy deployment.

Conformal Prediction (CP) \cite{vovk2005algorithmic,lei2018distribution} has emerged as a distribution-free framework that equips any black-box model with valid finite-sample coverage guarantees. However, directly adapting CP to human motion prediction poses significant challenges. 
\textbf{First}, the framework fundamentally relies on the assumption of \textit{exchangeability}, which is inherently violated by the strictly sequential and temporally correlated nature of human motion data.
\textbf{Second}, safety-critical motion prediction demands \textit{joint coverage} across complex, high-dimensional spatial-temporal configurations, but standard CP calibrates variables independently. Naively concatenating these dimension-wise bounds inevitably yields invalid or excessively inflated prediction regions.
\textbf{Third}, human motion is \textit{intrinsically structured}, but distribution-free calibration often yields uniform or structure-agnostic uncertainty regions. Constructing boundaries without integrating these structural priors leads to poorly conditioned uncertainty estimates, compromising the compactness of prediction regions.

To tackle these challenges, we propose \textbf{C}onformal \textbf{R}egions with \textbf{I}ntegrated \textbf{S}keletal \textbf{P}riors (CRISP), a model-agnostic conformal prediction framework tailored for uncertainty quantification in 3D human motion forecasting (as shown in Figure \ref{fig:1}). By constructing distribution-free prediction regions under any user-specified probability, CRISP seamlessly equips existing black-box predictors with the exact finite-sample mathematical reliability essential for safety-critical deployment. Specifically, inspired by time-series conformal prediction \cite{stankeviciute2021conformal}, CRISP treats entire motion sequences as \textit{i.i.d.} samples drawn from an underlying joint distribution, ensuring that any new test trajectory is strictly exchangeable with the calibration dataset. This trajectory-level exchangeability elegantly bypasses the strict point-wise \textit{i.i.d.} assumption, restoring finite-sample coverage guarantees. Building upon this, we move beyond standard isolated calibration to guarantee valid joint coverage across both multi-dimensional joints and multi-step horizons. To achieve this unified objective, CRISP introduces a joint spatial-temporal uncertainty quantification approach. \textbf{Spatially}, it projects uncertainty into covariance-parameterized ellipsoidal sets, dynamically aligning prediction boundaries with the complex geometry of human poses. \textbf{Temporally}, it bounds sequential trajectories via Copula functions \cite{sun2023copula}, seamlessly preserving this joint coverage guarantee over the entire forecasting horizon.

Furthermore, to effectively combine statistical calibration with rigid kinematics, CRISP introduces a topologically regularized covariance matrix. Our key insight is that human skeletal topology naturally restricts how uncertainty spreads across joints. Purely empirical statistics fail to distinguish these true kinematic dependencies from spurious correlations, inevitably leading to severe boundary inflation. Motivated by this, we develop a structured nonconformity score that fuses empirical residual covariance with the skeletal prior, producing the prediction regions that are mathematically reliable and highly compact.

Our contributions are summarized as follows:
\begin{itemize}
\item We propose an uncertainty quantification framework for 3D human motion forecasting, providing theoretical guarantees for reliable and trustworthy deployment. 
\item We achieve valid joint coverage across multi-dimensional joints and multi-step horizons by introducing covariance-parameterized ellipsoidal sets for spatial boundaries and Copula functions for temporal trajectories.
\item We inject the topology-aware prior as an inductive bias to actively regularize the uncertainty space. This allows the framework to capture intrinsic inter-joint dependencies, effectively producing highly compact prediction regions.
\item Extensive experiments on benchmarks demonstrate that CRISP consistently achieves valid target coverage with tighter prediction regions than representative baselines.
\end{itemize}
\section{Related Work}
\label{sec:related_work}

\begin{figure*}[t]
\centering
\includegraphics[width=\textwidth]{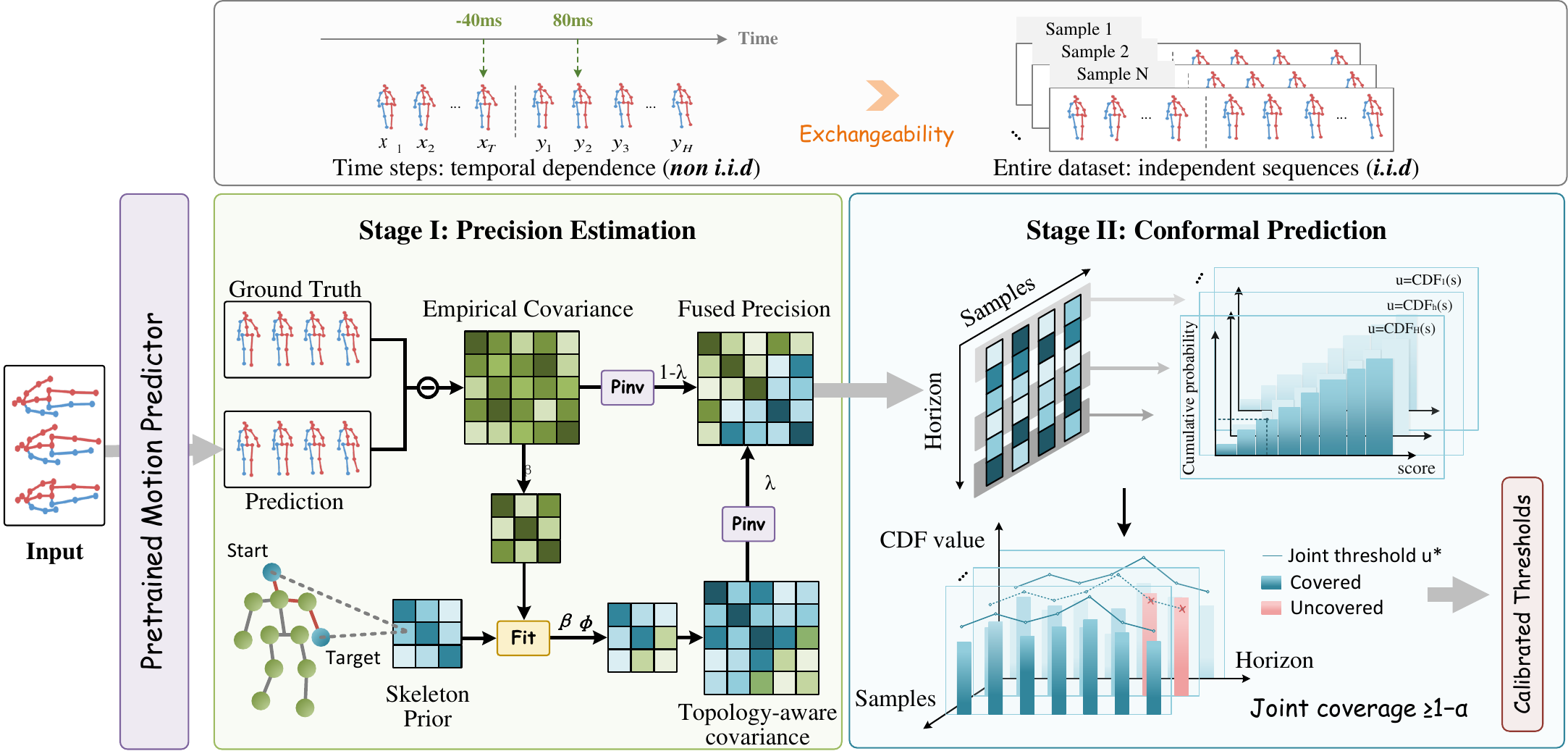} 
\caption{\textbf{Overview of our CRISP}. (Top) We establish sequence-level exchangeability by treating entire motion trajectories as i.i.d. samples. (Bottom) Spatially, empirical residual statistics are fused with a structural skeletal prior to estimate a topology-aware covariance, preventing isotropic inflation and structuring horizon-wise nonconformity scores. Temporally, an empirical copula calibrates cross-horizon dependencies into a joint threshold vector $\mathbf{u}^{\star}$, ensuring bounded multi-step prediction regions.} 
\vspace{-2mm}
\label{fig:fig2}
\end{figure*}

\textbf{Deterministic Motion Prediction.} 
The primary goal of deterministic prediction is to generate a single, highly accurate future pose sequence given historical observations. Early paradigms predominantly relied on recurrent architectures to autoregressively model temporal dynamics, such as ERD \cite{fragkiadaki2015recurrent} and Res-sup \cite{Res-sup}. To alleviate the error compounding issues in recurrent decoding, the field rapidly shifted towards feed-forward frameworks. A wide variety of architectures, such as Graph Convolutional Networks (GCNs) \cite{ltd,wang2024gcnext,zhou2026adaptive}, Transformers \cite{xu2023auxiliary,wei2025alien,tang2026human}, MLPs \cite{guo2023back,wei2025alien,wei2026progressive,hu2026breaking}, and Kolmogorov-Arnold Networks (KANs) \cite{KAN}, have been proposed to forecast entire sequences in a single pass. Despite these architectural advancements, deterministic paradigms fundamentally neglect the inherent uncertainty and multimodality naturally embedded in complex human motion patterns.

\noindent \textbf{Stochastic Motion Prediction.} 
To address the multimodality, stochastic approaches shift the objective from point estimation to distribution modeling. Mainstream methods leverage deep generative frameworks to map random latent noise into a set of plausible future trajectories. Early advancements primarily utilized Generative Adversarial Networks (GANs) \cite{zhao2023bidirectional} to enforce kinematic realism via adversarial training, and Variational Autoencoders (VAEs) \cite{yuan2020dlow,xu2024learning} to efficiently sample diverse hypotheses from structured latent spaces. Denoising Diffusion Probabilistic Models (DDPMs) \cite{chen2023humanmac,sun2024comusion} have emerged as a powerful alternative, employing denoising processes to achieve great generation fidelity and diversity. 
While these generative models excel at producing visually plausible outcomes, they output an unconstrained set of empirical hypotheses that fail to provide risk-controlled boundaries in practical deployment. 
More recently, SAGGB \cite{wang2024existence} and UnPOSed \cite{saadatnejad2024toward} have attempted to incorporate \textit{uncertainty awareness} into their modeling processes. However, these methods treat uncertainty merely as an auxiliary regression task by forcing complex kinematic dynamics into predefined Gaussian likelihoods under rigid parametric assumptions to output scalar variances. As these variances are empirically optimized rather than statistically calibrated, they fail to form coherent prediction regions and lack finite-sample theoretical guarantees for overall coverage. Departing from these empirical paradigms, we extend the distribution-free conformal calibration to reliable human motion prediction.

\noindent \textbf{Conformal Prediction.} 
Conformal Prediction (CP) \cite{vovk2005algorithmic,shafer2008tutorial} has emerged as a powerful statistical framework for rigorous uncertainty quantification across diverse fields, such as time-series forecasting \cite{auer2023conformal,sun2026conformal}, fine-grained vision-language models \cite{silva2025conformal,silva2025trustworthy}, long-tailed classification \cite{ding2026conformal}, and pose estimation \cite{zhang2025champ}. Its core strength is the ability to provide distribution-free, finite-sample coverage guarantees without strict parametric assumptions. For instance, ACI \cite{gibbs2021adaptive} dynamically adjusts to distribution shifts, while MultiDimSPCI \cite{xu2024conformal} utilizes quantile random forests but suffers from the computational burden of repeated model fitting. Crucially, these standard methods necessitate the assumption of data exchangeability, which is inherently violated by sequential data. To circumvent this, recent time-series CP variants have explored sequential quantile regression and Copula-based dependency modeling. CF-RNN \cite{stankeviciute2021conformal} constructs horizon-wise intervals with a Bonferroni correction, which can become overly conservative as the forecasting horizon increases. Instead, CopulaCPTS \cite{sun2023copula} models dependence across horizons using a copula function, achieving more efficient joint calibration. 
However, when applied to high-dimensional human kinematics, these generic time-series adaptations fundamentally overlook the intrinsic spatial dependencies of the human skeletal topology, yielding conservative and inflated prediction boundaries. 
\section{Proposed Method}
\subsection{Problem Setup \& Overview}
\label{sec:problem}

\noindent \textbf{Motivation.}
Let $\mathbf{X} = \{\mathbf{x}_1, \mathbf{x}_2, \cdots, \mathbf{x}_{T}\}$ denote the history of $T$ frames, where each pose $\mathbf{x}_t = [\mathbf{x}_{t,1}, \mathbf{x}_{t,2}, \cdots, \mathbf{x}_{t,J}]\in\mathbb{R}^{J\times 3}$ contains the 3D coordinates of $J$ body joints, and $\mathbf{x}_{t,j}$ corresponds to the position of joint $j$ at time $t$. 
Human motion prediction typically aims to predict the subsequent $H$ future poses $\widehat{\mathbf{Y}}=\{\widehat{\mathbf{y}}_{1},\widehat{\mathbf{y}}_{2},\cdots,\widehat{\mathbf{y}}_{H}\}$ as accurately as possible towards the ground truth $\mathbf{Y}$.
Existing motion prediction methods focus on designing a dedicated network architecture $f_{\theta}(\cdot)$ to capture spatial-temporal dependencies:
\begin{equation}
\min_{\theta}\|f_{\theta}(\mathbf{X})-\mathbf{Y}\|_{2}^{2}.
\end{equation}
However, human motion is inherently uncertain, as a single past observation can naturally unfold into various valid futures.
Relying only on a single deterministic future sequence fails to convey predictive uncertainty. Without quantifying this underlying uncertainty, 
it is particularly risky to deploy such predictors in applications such as autonomous driving and human–robot interaction, where decision-making must account for potential risks.

\noindent \textbf{Problem Setup.}
To mitigate these risks, rather than relying on a single point estimate, we aim to quantify the predictive uncertainty by constructing a statistical confidence region. Formally, given a fixed pretrained predictor $f_{\theta}(\cdot)$ and a user-specified target confidence level $1-\alpha\in (0,1)$, our goal is to construct step-wise prediction regions $\Gamma_t^{1-\alpha}(\mathbf{X})$ from an observed sequence, such that the ground-truth is simultaneously covered across all forecasting horizons:
\begin{equation}
\mathbb{P}
\left[
    \forall t\in\{1,2,\cdots,H\},\;
    \mathbf{y}_t\in \Gamma_t^{1-\alpha}(\mathbf{X})
\right]
\geq 1-\alpha.
\end{equation}
To achieve this, we assume access to a pool of data disjoint from both the training set $\mathcal{D}_{\mathrm{train}}$ and the evaluation set $\mathcal{D}_{\mathrm{test}}$. Specifically, we utilize an estimation set $\mathcal{D}_{\mathrm{est}}$ to compute the empirical residual covariance. The remaining calibration data is partitioned into two independent subsets: $\mathcal{D}_{\mathrm{cal}-1}$ for estimating marginal CDFs, and $\mathcal{D}_{\mathrm{cal}-2}$ for conducting copula-based joint calibration.

\noindent \textbf{CRISP Overview.} 
Figure~\ref{fig:fig2} illustrates the CRISP framework. Assuming sequence-level exchangeability, CRISP constructs reliable prediction regions through two synergistic steps. 
Spatially, by exploiting the inherent kinematic correlations among human joints, it integrates empirical residual statistics with a skeletal prior to estimate a topology-aware covariance, preventing isotropic inflation and shaping structurally aware nonconformity scores. 
Temporally, it calibrates the cross-horizon dependence of these scores via an empirical copula to derive a joint threshold vector $\mathbf{u}^{\star}$. In essence, the topology-aware covariance characterizes the spatial geometry of pose deviations, while the copula calibration dictates their joint bounds across forecasting horizons, yielding compact prediction regions.

\subsection{Precision Estimation with Skeletal Priors}
\noindent \textbf{Exchangeability Setting.} 
Inspired by time series conformal prediction \cite{stankeviciute2021conformal}, we regard each complete motion sequence as an exchangeable sample, allowing arbitrary spatial and temporal dependencies within the trajectory. As depicted in Figure \ref{fig:fig2}, we have the following assumption.
\begin{assumption}[Sequence-level exchangeability]
\label{ass:exchangeability}
The motion sequences in
$\mathcal D_{\mathrm{cal}\text{-}1}
\cup\mathcal D_{\mathrm{cal}\text{-}2}$
and a new test sequence are exchangeable. Arbitrary dependence is
allowed among body joints and forecasting horizons within each
sequence.
\end{assumption}

\label{sec:topology_covariance}
\noindent \textbf{Skeletal Prior and Empirical Covariance.}
As shown in Figure~\ref{fig:viz}, human motion exhibits heterogeneous spatial uncertainty: central joints (e.g., torso) typically present minimal predictive variance, whereas distal extremities show significantly larger kinematic deviations. Moreover, kinematic coupling through the skeleton can give rise to correlated prediction errors across joints.
Empirical residual covariance can capture these dependencies, but its estimation is subject to finite-sample error, which can be amplified by inversion. Inspired by \cite{zhang2026topology}, we leverage human skeletal topology as an inductive prior to regularize the empirical residual covariance for better modeling of these structural correlations. Specifically, we first estimate the residual covariance from $\mathcal{D}_{\mathrm{est}}$, construct a topology-induced covariance from the skeletal graph, and then fuse the two estimates.
For the $i$-th sample, the vectorized residual of each timestamp $t\in \{1,2,\cdots,H\}$ is defined as 
$ \mathbf{r}_t^i=\operatorname{vec}(\mathbf{y}_t^i-\widehat{\mathbf{y}}_t^i)\in\mathbb R^{3J}$. With $\bar{\mathbf{r}}$ as the sample mean, the empirical covariance $\widehat{\mathbf{\Sigma}}_{\mathrm{emp}}$ is then estimated as:
\begin{equation}
\begin{aligned}
\widehat{\mathbf{\Sigma}}_{\mathrm{emp}}
&=
\frac{1}{N_{\mathrm{est}}{\times}H-1}
\sum_{i=1}^{N_{\mathrm{est}}}
\sum_{t=1}^{H}
(\mathbf{r}_t^i-\bar{\mathbf{r}})
(\mathbf{r}_t^i-\bar{\mathbf{r}})^\top,
\end{aligned}
\label{eq:sample_covariance}
\end{equation}
where $N_{\mathrm{est}} = \vert{}\mathcal{D}_{\mathrm{est}}\vert{}$. 
The empirical covariance captures variations in prediction errors across joints and their correlations, providing a characterization of spatial uncertainty.

We further introduce a structural prior through the skeletal graph $\mathcal{G}=(\mathcal{V},\mathcal{E})$, where $\mathcal{V}$ denotes the set of body joints and $\mathcal{E}$ denotes their anatomical connections. Let $D_{ab}$ denote the shortest-path distance between joints $a$ and $b$ along the edges of the skeleton graph $\mathcal{G}$. 
We model the topology covariance using an exponential kernel \(K(\beta,\phi)\). Here, $\beta$ determines the covariance scale and $\phi$ controls the decay length along the skeleton. For any joint pair \((a,b)\), we have:
\begin{equation}
K_{ab}^{\mathrm{topo}}(\beta,\phi)
=\beta\exp\left(-D_{ab}/\phi\right).
\label{eq:topo_kernel}
\end{equation}

Since skeletal distances alone do not determine the scale and decay of residual covariance, the kernel parameters are fitted to the empirical estimates.
To reconcile the joint-level kernel with the coordinate-level empirical covariance, we represent each empirical covariance block by the mean of its diagonal entries, denoted by $K_{ab}^{\mathrm{emp}}$. The kernel parameters are obtained by least-squares fitting:
\begin{equation}
(\beta^\ast,\phi^\ast)=
\arg\min_{\beta,\phi}
\sum_{(a,b):D_{ab}<\infty}
\left[
K_{ab}^{\mathrm{emp}}-
K_{ab}^{\mathrm{topo}}(\beta,\phi)
\right]^2.
\label{eq:fit_para}
\end{equation}
The fitted covariance is extended to the coordinate level via the Kronecker product.
To balance data-driven dependence with skeletal prior information, we define the fused precision matrix ${\mathbf{A}}_{\mathrm{\lambda}}$ by linearly combining the pseudo-inverses of empirical and topology-induced covariance estimates:
\begin{equation}
\mathbf{A}_\mathrm{\lambda}=
(1-\lambda)\widehat{\mathbf{\mathbf{\Sigma}}}_{\mathrm{emp}}^{-1}+
\lambda\widehat{\mathbf{\mathbf{\Sigma}}}_{\mathrm{topo}}^{-1}.
\label{eq:fused_covariance}
\end{equation}
Here, $\lambda\in[0,1]$ controls the relative contributions of the two precision estimates. The resulting precision matrix $\mathbf{A}_\mathrm{\lambda}$ provides a more robust covariance estimation, and is used to define the nonconformity score below.

\subsection{Cross-Horizon Calibration}
\label{sec:copula_calibration}
\noindent \textbf{Topology-aware Nonconformity Score.}
For each calibration sample
\(
(\mathbf{X}^{i},\mathbf{Y}^{i})
\in\mathcal{D}_{\mathrm{cal}-1}\cup\mathcal{D}_{\mathrm{cal}-2}
\), 
we compute a nonconformity score \(s_t^i\) at each horizon \(t\).
We define the topology-aware score using the Mahalanobis distance, that accounts for residual scales and cross-joint correlations:
\begin{equation}
s_t(\mathbf{y}_{t},\widehat{\mathbf{y}_{t}})=
(\mathbf{r}_{t}-\bar{\mathbf{r}})^{\top}
{\mathbf{A}}_{\mathrm{\lambda}}
(\mathbf{r}_{t}-\bar{\mathbf{r}}).
\label{eq:nonconformity_score}
\end{equation}

\noindent \textbf{Marginal CDF Estimation.}
Given that score distributions may vary across forecasting horizons, we first map the horizon-wise scores to a common probability scale. 
Specifically, the first subset \(\mathcal D_{\mathrm{cal}-1}\) is used to estimate marginal cumulative distribution at each horizon.
\begin{equation}
\widehat{F}_t(s)=
\frac{1}{|\mathcal{D}_{\mathrm{cal}\text{-}1}|+1}
\sum_{i\in\mathcal{D}_{\mathrm{cal}\text{-}1}}
\mathbb{I}\{s_t^i<s\}.
\label{eq:marginal_cdf}
\end{equation}
\(\widehat F_t(s)\) represents a rank-based cumulative score estimate at horizon \(t\), indicating the relative position of \(s\) among the corresponding calibration scores.
As shown in Figure~\ref{fig:fig2}, horizontal gaps reflect score differences, while each observation contributes equally to the cumulative estimate.
With these mappings fixed, we transform each sequence $i\in\mathcal{D}_{\mathrm{cal}\text{-}2}$ into $\mathbf{u}^i=(u_1^i,\ldots,u_H^i)$, where $u_t^i=\widehat{F}_t(s_t^i).$
These vectors serve as the input to copula calibration.

\begin{algorithm}[tb]
\caption{Motion Conformal Prediction (CRISP)}
\label{alg:CRISP}
\textbf{Input}: 
    estimation set $\mathcal{D}_{\mathrm{est}}$,
    calibration dataset $\mathcal{D}_{\mathrm{cal}\text{-}1}$ and $\mathcal{D}_{\mathrm{cal}\text{-}2}$,
    skeleton graph $\mathcal{G}$,
    pretrained predictor$f_{\theta}$,
    test observation $\mathbf{X}$,
    fusion weight $\lambda$,
    and the target level $1-\alpha$.\\
\textbf{Output}:
    ${\Gamma}_{1}^{1-\alpha}$, 
    ${\Gamma}_{2}^{1-\alpha}$,
    $\cdots$,
    ${\Gamma}_{H}^{1-\alpha}$ for test data.
\begin{algorithmic}[1]
    \For{each sample $i$ in $\mathcal D_{\mathrm{est}}$}
    \State Compute the residuals
    $\{\mathbf{r}_t^i\}_{t=1}^{H}$ using $f_{\theta}$.
    \EndFor
    \State Estimate
    $\widehat{\mathbf{\mathbf{\Sigma}}}_{\mathrm{emp}}$
    from the residuals using Eq.~\eqref{eq:sample_covariance}.
    \State Compute the shortest-path distance matrix
    $\{D_{ab}\}_{a,b=1}^{J}$ from the skeleton graph $\mathcal{G}$.
    \State Construct
    $\widehat{\mathbf{\mathbf{\Sigma}}}_{\mathrm{topo}}$ using Eqs.~\eqref{eq:topo_kernel} and~\eqref{eq:fit_para}.
    \State Obtain the fused matrix $ 
    {\mathbf{A}}_{\mathrm{\lambda}}$
    using Eq.~\eqref{eq:fused_covariance}.

    \For{each sample $i$ in
    $\mathcal{D}_{{cal}\text{-}1}\cup\mathcal{D}_{\mathrm{{cal}\text{-}2}}$}
        \State
        compute $s_t^i$ for {$t\in\{1,2,\cdots,H\}$} using Eq.~\eqref{eq:nonconformity_score}.
    \EndFor
    \State Compute $\widehat{F}_t(s_t)$ on $\mathcal{D}_{\mathrm{cal}\text{-}1}$ by Eq.~\eqref{eq:marginal_cdf}, $t=1,\ldots,H$.
    \State Construct empirical Copula $C_{\mathrm{emp}}$ on $\mathcal{D}_{\mathrm{cal}\text{-}2}$ by Eq.~\eqref{eq:empirical_copula}.
    \State Find
    $\mathbf{u}^\star$ using~\eqref{eq:copula_threshold}.
    \State Compute $s_t^\star=\widehat{F}_t^{-1}(u_t^\star)$ for {$t\in\{1,2,\cdots,H\}$}.
    \State Construct
    $\Gamma_t^{1-\alpha}(\mathbf{X})=\{\mathbf{y}:s_t(\mathbf{y},\widehat {\mathbf{y}}_t)\leq s_t^\star\}$
    for $t\in\{1,2,\cdots,H\}$.
    \State \textbf{return}
    $\{\Gamma_t^{1-\alpha}(\mathbf{X})\}_{t=1}^{H}$.
\end{algorithmic}
\end{algorithm}

\noindent \textbf{Copula Calibration.}
To account for temporal dependencies across prediction horizons, we construct an empirical copula from the resulting vectors \(\mathbf{u}^i=(u_1^i,\cdots,u_H^i)\)
\begin{equation}
C_{\mathrm{emp}}(\mathbf{u})=
\frac{1}{|\mathcal{D}_{\mathrm{cal}-2}|+1}
\sum_{i\in\mathcal{D}_{\mathrm{cal}-2}\cup\{\infty\}}
\prod_{t=1}^{H}
\mathbb{I}\{u_t^i \leq u_t\}.
\label{eq:empirical_copula}
\end{equation}
Here, $\mathbf{u}^{\infty}=(+\infty,\ldots,+\infty)$ is an $H$-dimensional vector, where $\infty$ denotes an additional index. We aim to search for the threshold vector by solving
\begin{equation}
\boldsymbol{{u}}^{\star}
\in
\arg\min_{\mathbf{u}\in[0,1]^H}
\sum_{t=1}^{H}{u_t}
\quad
\mathrm{s.t.} 
\quad
C_{\mathrm{emp}}(\boldsymbol{\mathbf{u}})
\geq 1-\alpha.
\label{eq:copula_threshold}
\end{equation}
The calibrated nonconformity threshold for timestamp $t$ is 
$s_t^\star=\widehat F_t^{-1}({u}_t^\star)$. 
Collectively, the regions $\{\Gamma_t^{1-\alpha}(\mathbf{X})\}_{t=1}^{H}$ form the full-horizon prediction region; a future sequence is covered if and only if $\mathbf{y}_{t}\in\Gamma_t^{1-\alpha}(\mathbf{X})$ for every time step $t$. The overall procedure of our proposed CRISP framework is summarized in Algorithm~\ref{alg:CRISP}.

\subsection{Theoretical Analysis}
\label{sec:theory}
We establish a finite-sample simultaneous coverage guarantee for the multivariate, multi-step motion forecasts from CRISP. The complete proof is provided in the Appendix.
\begin{theorem}[Validity of CRISP]
Under Assumption 3.1, CRISP produces valid prediction regions across the entire forecast horizon. \ie
\begin{equation}
\mathbb P
\left[
\forall t\in\{1,2,\cdots,H\},\;
\mathbf{y}_t\in\Gamma_t^{1-\alpha}(\mathbf{X})
\right]
\geq 1-\alpha.
\label{eq:full_horizon_validity}
\end{equation}
\end{theorem}
\section{Experiments}
\label{sec:exp}

\begin{table*}[t]
\centering
    \caption{Comparison of CRISP against baselines on Human3.6M, utilizing EqMotion (long-term) as the underlying predictor. The target confidence level is set to 90\%. Efficiency is reported as mean $\pm$ standard deviation, with the best results highlighted in \textbf{bold}.}
    \label{tab:eqmotion_h36m_long}
\vspace{-2mm} 
\footnotesize 
\begin{tblr}
{
    width   = \linewidth,
    colsep  = 2pt,
    colspec = {c *{4}{X[0.7,c] X[1.3,c]}},
    cells = {halign=c, valign=m},
    stretch = 0, 
    rowsep = 1pt,
    column{1} = {halign=l, leftsep=5pt},
    vline{2,4,6,8} = {0.25pt, abovepos=-1, belowpos=-1},
    hborder{1-3,8-10,15-17,22-24,Z} = {abovespace=1.5pt, belowspace=1.5pt},
}

\hline[0.8pt]
    Action
    & \SetCell[c=2]{c,m} Walking &
    & \SetCell[c=2]{c,m} Eating &
    & \SetCell[c=2]{c,m} Smoking &
    & \SetCell[c=2]{c,m} Discussion & \\
\hline
    Method
    & Coverage & Efficiency $\downarrow$ 
    & Coverage & Efficiency $\downarrow$ 
    & Coverage & Efficiency $\downarrow$ 
    & Coverage & Efficiency $\downarrow$ \\
\hline
    MultiDimSPCI~\cite{xu2024conformal}
        & 90.94 & 153.07 $\pm$ 8.46 
        & 88.85 & \textbf{118.52 $\boldsymbol{\pm}$ 4.46}
        & 90.76 & 140.21 $\pm$ 2.92 
        & 88.50 & 162.72 $\pm$ 6.81 \\
    CopulaCPTS~\cite{sun2023copula}
        & 91.57 & 311.56 $\pm$ 24.74 
        & 88.98 & 364.25 $\pm$ 12.41
        & 89.96 & 334.66 $\pm$ 8.59 
        & 90.04 & 509.13 $\pm$ 14.65 \\
    CF-RNN~\cite{stankeviciute2021conformal}
        & 98.92 & 436.97 $\pm$ 5.15 
        & 98.12 & 467.38 $\pm$ 30.82
        & 98.13 & 413.76 $\pm$ 8.60 
        & 98.14 & 631.49 $\pm$ 19.28 \\
    Shuffle-Cov
        & 90.87 & \textbf{135.03 $\boldsymbol{\pm}$ 4.65} 
        & 89.62 & 261.22 $\pm$ 29.12
        & 90.73 & 256.00 $\pm$ 28.46 
        & 89.99 & 335.92 $\pm$ 35.09 \\
    \textbf{CRISP (Ours)}
        & 91.02 & 135.63 $\pm$ 5.65 
        & 88.59 & 119.04 $\pm$ 3.83
        & 91.07 & \textbf{132.21 $\boldsymbol{\pm}$ 3.82} 
        & 88.74 & \textbf{148.96 $\boldsymbol{\pm}$ 5.07} \\
\hline
    Action
    & \SetCell[c=2]{c,m} Directions &
    & \SetCell[c=2]{c,m} Greeting &
    & \SetCell[c=2]{c,m} Phoning &
    & \SetCell[c=2]{c,m} Posing & \\
\hline
    Method
    & Coverage & Efficiency $\downarrow$ 
    & Coverage & Efficiency $\downarrow$ 
    & Coverage & Efficiency $\downarrow$ 
    & Coverage & Efficiency $\downarrow$ \\
\hline
    MultiDimSPCI~\cite{xu2024conformal}
        & 90.71 & 182.66 $\pm$ 4.62 
        & 89.56 & 215.95 $\pm$ 1.85
        & 89.69 & 166.79 $\pm$ 4.59 
        & 89.66 & 239.04 $\pm$ 5.13 \\
    CopulaCPTS~\cite{sun2023copula}
        & 90.43 & 547.37 $\pm$ 15.21 
        & 88.98 & 756.25 $\pm$ 13.24
        & 88.85 & 461.12 $\pm$ 21.56 
        & 90.94 & 660.65 $\pm$ 15.88 \\
    CF-RNN~\cite{stankeviciute2021conformal}
        & 97.69 & 649.26 $\pm$ 10.76 
        & 97.51 & 893.21 $\pm$ 13.51
        & 97.20 & 593.79 $\pm$ 7.73 
        & 98.37 & 838.13 $\pm$ 20.13 \\
    Shuffle-Cov
        & 90.35 & 319.60 $\pm$ 46.31 
        & 88.87 & 508.58 $\pm$ 70.32
        & 89.14 & 351.57 $\pm$ 18.87 
        & 90.75 & 417.46 $\pm$ 41.12 \\
    \textbf{CRISP (Ours)}
        & 90.53 & \textbf{164.10 $\boldsymbol{\pm}$ 5.75} 
        & 89.75 & \textbf{212.36 $\boldsymbol{\pm}$ 1.58}
        & 89.62 & \textbf{165.61 $\boldsymbol{\pm}$ 3.68} 
        & 90.04 & \textbf{216.49 $\boldsymbol{\pm}$ 5.41} \\
\hline
    Action
    & \SetCell[c=2]{c,m} Purchases &
    & \SetCell[c=2]{c,m} Sitting &
    & \SetCell[c=2]{c,m} Sitting Down &
    & \SetCell[c=2]{c,m} Taking Photo & \\
\hline
    Method
    & Coverage & Efficiency $\downarrow$ 
    & Coverage & Efficiency $\downarrow$ 
    & Coverage & Efficiency $\downarrow$ 
    & Coverage & Efficiency $\downarrow$ \\
\hline
    MultiDimSPCI~\cite{xu2024conformal}
        & 90.52 & 224.69 $\pm$ 12.68 
        & 90.37 & 219.05 $\pm$ 6.01
        & 91.67 & 299.11 $\pm$ 6.48 
        & 90.84 & 218.88 $\pm$ 11.48 \\
    CopulaCPTS~\cite{sun2023copula}
        & 91.93 & 670.97 $\pm$ 14.76 
        & 89.57 & 502.29 $\pm$ 18.48
        & 90.54 & 715.16 $\pm$ 23.37 
        & 89.99 & 519.64 $\pm$ 11.66 \\
    CF-RNN~\cite{stankeviciute2021conformal}
        & 97.87 & 834.02 $\pm$ 13.72 
        & 97.86 & 571.57 $\pm$ 12.71
        & 98.27 & 934.08 $\pm$ 29.65 
        & 97.74 & 628.00 $\pm$ 12.01 \\
    Shuffle-Cov
        & 91.82 & 523.04 $\pm$ 41.61 
        & 90.28 & 374.41 $\pm$ 21.75
        & 91.00 & 555.33 $\pm$ 47.49 
        & 90.07 & 410.50 $\pm$ 24.07 \\
    \textbf{CRISP (Ours)}
        & 90.91 & \textbf{220.62 $\boldsymbol{\pm}$ 6.40} 
        & 90.65 & \textbf{204.47 $\boldsymbol{\pm}$ 5.52}
        & 91.62 & \textbf{268.97 $\boldsymbol{\pm}$ 5.85} 
        & 90.83 & \textbf{204.05 $\boldsymbol{\pm}$ 10.41} \\
\hline
    Action
    & \SetCell[c=2]{c,m} Waiting &
    & \SetCell[c=2]{c,m} Walking Dog &
    & \SetCell[c=2]{c,m} Walking Together &
    & \SetCell[c=2]{c,m} Average & \\
\hline
    Method
    & Coverage & Efficiency $\downarrow$ 
    & Coverage & Efficiency $\downarrow$ 
    & Coverage & Efficiency $\downarrow$ 
    & Coverage & Efficiency $\downarrow$ \\
\hline
    MultiDimSPCI~\cite{xu2024conformal}
        & 89.77 & 166.88 $\pm$ 5.49 
        & 90.41 & 256.36 $\pm$ 6.90
        & 90.71 & 163.28 $\pm$ 2.50 
        & 90.20 & 195.15 $\pm$ 1.98 \\
    CopulaCPTS~\cite{sun2023copula}
        & 90.04 & 524.70 $\pm$ 21.64 
        & 90.30 & 784.25 $\pm$ 28.01
        & 89.89 & 298.52 $\pm$ 6.97 
        & 90.13 & 530.70 $\pm$ 4.99 \\
    CF-RNN~\cite{stankeviciute2021conformal}
        & 98.05 & 688.14 $\pm$ 36.39 
        & 97.63 & 982.30 $\pm$ 13.54
        & 98.47 & 401.61 $\pm$ 27.84 
        & 98.00 & 664.25 $\pm$ 3.19 \\
    Shuffle-Cov
        & 90.22 & 386.50 $\pm$ 33.72 
        & 90.20 & 604.77 $\pm$ 70.23
        & 90.13 & 188.26 $\pm$ 13.06 
        & 90.27 & 375.21 $\pm$ 29.21 \\
    \textbf{CRISP (Ours)}
        & 89.92 & \textbf{165.55 $\boldsymbol{\pm}$ 4.04} 
        & 90.41 & \textbf{244.60 $\boldsymbol{\pm}$ 5.84}
        & 90.95 & \textbf{146.50 $\boldsymbol{\pm}$ 3.27} 
        & 90.31 & \textbf{183.28 $\boldsymbol{\pm}$ 2.10} \\
\hline[0.8pt]
\end{tblr}
\vspace{-3mm}
\end{table*}

\subsection{Datasets}
\textbf{Human3.6M} \cite{ionescu2013human3} is a large-scale dataset containing motion sequences from 7 subjects across 15 actions.
The original exponential map representations are converted into 3D joint coordinates, with motion sequences downsampled to 25 Hz.
We use S5 for uncertainty calibration and evaluation.
S11 is separately used for covariance estimation.

\noindent\textbf{AMASS} \cite{mahmood2019amass} collects multiple Mocap datasets under the SMPL representation. Following \cite{mao2020history,guo2023back}, we downsample the sequences to 25 fps and retain 18 body joints.
BMLrub is held out to evaluate uncertainty quantification,
while the remaining subsets are used to train the base predictor.

\noindent\textbf{3DPW} \cite{von2018recovering} is a challenging dataset containing motion sequences captured in indoor and outdoor scenes. 
Similar to \cite{mao2020history}, we apply the AMASS-trained model to 3DPW. 
For covariance estimation, we use the official validation split and reserve the test split exclusively for uncertainty evaluation.

\noindent\textbf{CHICO}~\cite{sampieri2022pose} is an industrial human-robot collaboration dataset covering 7 assembly-line motions from 20 operators. Following \citep{sampieri2022pose}, we retain 15 joints and use subjects 2, 3, 18, and 19 for uncertainty evaluation.
Additionally, CHICO includes 226 recorded human-robot collisions.

\subsection{Baselines \& Implementation Details}
\textbf{Baselines.} We compare the performance of our model with representative conformal prediction baselines: CF-RNN \cite{stankeviciute2021conformal}, CopulaCPTS \cite{sun2023copula}, and MultiDimSPCI \cite{xu2024conformal}. Additionally, we introduce an ablation variant, denoted as Shuffle-Cov, which randomly permutes the joint labels of the skeletal distance matrix prior to refitting the topology-induced covariance. To ensure a fair comparison, all methods are instantiated with the same state-of-the-art motion predictors under identical calibration settings.

\begin{figure}[t]
\centering
\includegraphics[width=1.0\columnwidth]{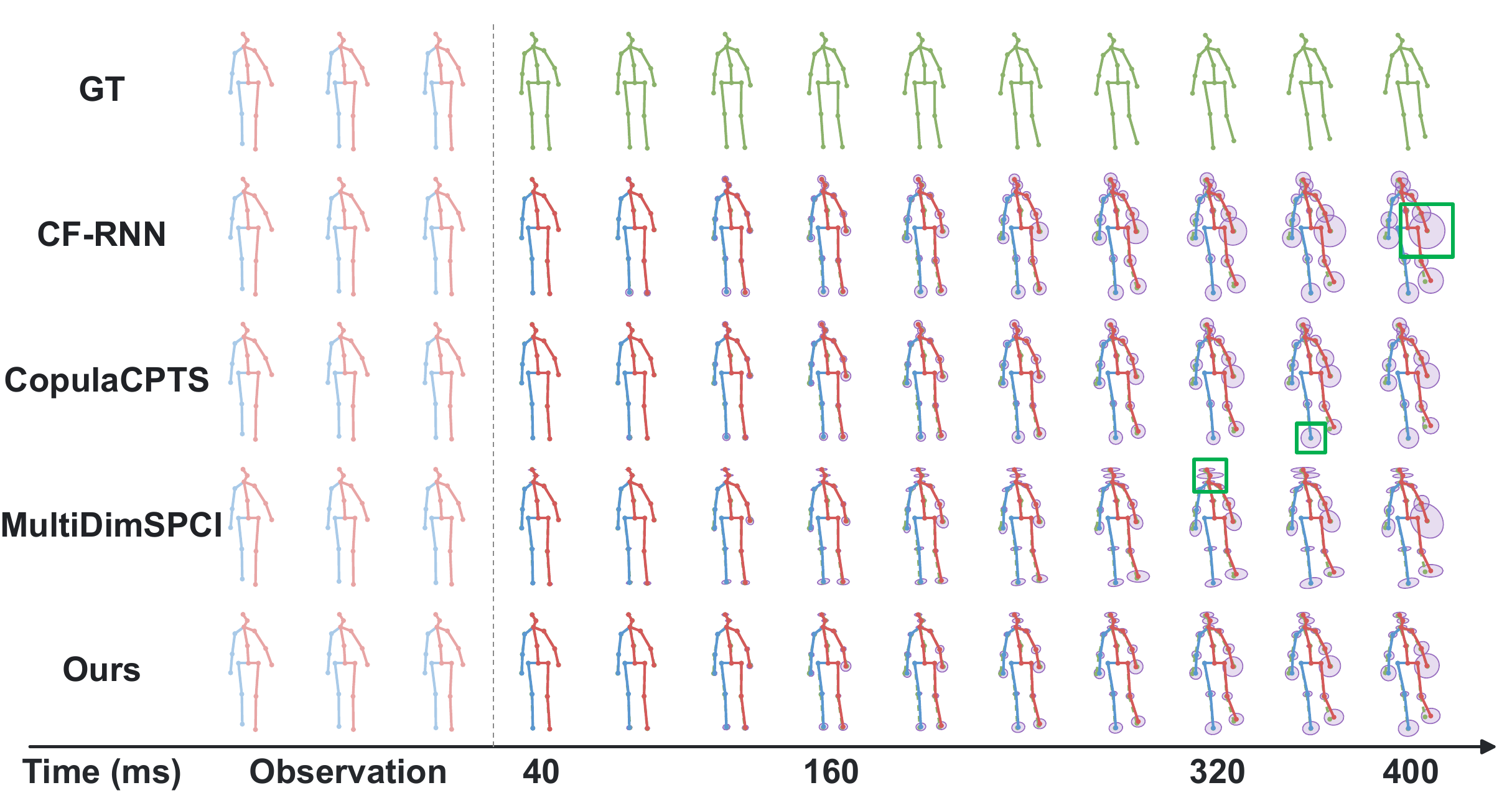} 
\vspace{-5mm}
\caption{Visualization of calibrated uncertainty regions on a ``Walking'' sequence. Projected joint-wise boundaries are depicted as purple ellipses alongside the GT poses (green).}
\vspace{-4mm}
\label{fig:viz}
\end{figure}

\noindent \textbf{Implementation Details.} 
We instantiate our framework using PGBIG ~\cite{pgbig}, EqMotion ~\cite{xu2023eqmotion}, siMLPe ~\cite{guo2023back}, and SeS-GCN ~\cite{sampieri2022pose}.
Following ~\cite{pgbig}, we utilize 10 observed frames to predict 10 (400 ms) or 25 (1,000 ms) future frames, keeping all pretrained forecasters frozen during calibration. 
The target coverage $1- \alpha$ and the fusion weight $\lambda$ are set to $90\%$ and $0.6$, respectively, with the topology-kernel parameters fitted via least squares.
To enable gradient-based optimization, we approximate the indicator functions with sigmoids. 
The threshold vector $\boldsymbol{u} \in [0,1]^H$ (initialized to ones) is optimized via Adam for 800 iterations (learning rate $10^{-3}$, weight decay $10^{-4}$).
  All reported metrics are presented as the mean and standard deviation computed across five independent runs. More details can be found in the Appendix.

\begin{table*}[t]
\centering
\caption{Average coverage and efficiency on Human3.6M using PGBIG (long-term) and siMLPe (long-term) with and without TCL, and EqMotion (short-term) as base predictors. The target coverage is 90\%. \textbf{Bold} indicates the lowest mean efficiency for each predictor.}
\label{tab:alternative_predictors}
\vspace{-2mm}
\footnotesize
\resizebox{\linewidth}{!}{%
\begin{tblr}
{
    colspec = {l *{5}{cc}},
    colsep  = 1pt,
    stretch = 0,
    rowsep  = 1pt,
    cells   = {halign=c, valign=m},
    column{1} = {halign=l, leftsep=5pt},
    vline{2,4,6,8,10} = {
        0.25pt,
        abovepos=-1,
        belowpos=-1
    },
    hborder{1-3,Z} = {abovespace=1.5pt, belowspace=1.5pt},
}

\hline[0.8pt]
    Predictor
    & \SetCell[c=2]{c,m} PGBIG &
    & \SetCell[c=2]{c,m} PGBIG + TCL &
    & \SetCell[c=2]{c,m} siMLPe &
    & \SetCell[c=2]{c,m} siMLPe + TCL &
    & \SetCell[c=2]{c,m} EqMotion & \\
\hline
    Method
    & Coverage & Efficiency $\downarrow$
    & Coverage & Efficiency $\downarrow$
    & Coverage & Efficiency $\downarrow$
    & Coverage & Efficiency $\downarrow$
    & Coverage & Efficiency $\downarrow$ \\
\hline
    MultiDimSPCI~\cite{xu2024conformal}
        & 90.24 & 219.52 $\pm$ 0.99
        & 90.23 & 218.74 $\pm$ 1.22
        & 90.29 & 251.00 $\pm$ 2.39
        & 90.12 & 257.08 $\pm$ 1.23
        & 90.17 & 112.75 $\pm$ 1.00 \\
    CopulaCPTS~\cite{sun2023copula}
        & 90.13 & 550.05 $\pm$ 7.37
        & 90.00 & 544.55 $\pm$ 6.86
        & 90.07 & 560.52 $\pm$ 2.68
        & 90.04 & 564.21 $\pm$ 5.39
        & 90.30 & 254.80 $\pm$ 3.53 \\
    CF-RNN~\cite{stankeviciute2021conformal}
        & 97.81 & 680.56 $\pm$ 4.87
        & 97.85 & 677.89 $\pm$ 3.32
        & 97.92 & 701.25 $\pm$ 5.91
        & 97.99 & 712.73 $\pm$ 9.47
        & 96.59 & 304.24 $\pm$ 0.84 \\
    Shuffle-Cov
        & 90.27 & 387.97 $\pm$ 23.57
        & 90.10 & 382.98 $\pm$ 24.57
        & 90.21 & 387.05 $\pm$ 24.80
        & 90.38 & 386.90 $\pm$ 27.03
        & 90.32 & 193.35 $\pm$ 13.29 \\
    \textbf{CRISP (Ours)}
        & 90.19 & \textbf{187.27 $\boldsymbol{\pm}$ 0.92}
        & 90.09 & \textbf{185.38 $\boldsymbol{\pm}$ 1.17}
        & 90.30 & \textbf{204.55 $\boldsymbol{\pm}$ 2.52}
        & 90.29 & \textbf{209.46 $\boldsymbol{\pm}$ 1.59}
        & 90.04 & \textbf{108.13 $\boldsymbol{\pm}$ 1.03} \\
\hline[0.8pt]
\end{tblr}}
\vspace{-3mm}
\end{table*}

\begin{table*}[t]
\centering
\caption{Comparison with baselines on CHICO, utilizing SeS-GCN (short-term) as the predictor. The target confidence level is set to 90\%. 
}
\label{tab:chico_normal_short}
\vspace{-2mm}
\footnotesize
\begin{tblr}
{
    width   = \linewidth,
    colsep  = 1pt,
    colspec = {c *{4}{X[0.7,c] X[1.3,c]}},
    cells   = {halign=c, valign=m},
    stretch = 0,
    rowsep  = 1pt,
    column{1} = {halign=l, leftsep=5pt},
    vline{2,4,6,8} = {
        0.25pt,
        abovepos=-1,
        belowpos=-1
    },
    hborder{1-3,Z} = {
        abovespace=1.5pt,
        belowspace=1.5pt,
    }
}
\hline[0.8pt]
    Action
    & \SetCell[c=2]{c,m} Hammer &
    & \SetCell[c=2]{c,m} High Lift &
    & \SetCell[c=2]{c,m} Prec. P\&P &
    & \SetCell[c=2]{c,m} Rnd. P\&P & \\
\hline
    Method
    & Coverage & Efficiency $\downarrow$
    & Coverage & Efficiency $\downarrow$
    & Coverage & Efficiency $\downarrow$
    & Coverage & Efficiency $\downarrow$ \\
\hline
    MultiDimSPCI~\cite{xu2024conformal}
        & 90.35 & 128.58 $\pm$ 6.70
        & 89.99 & 156.10 $\pm$ 4.29
        & 90.20 & 99.09 $\pm$ 1.76
        & 89.01 & 110.95 $\pm$ 3.58 \\
    CopulaCPTS~\cite{sun2023copula}
        & 89.83 & 178.13 $\pm$ 3.67
        & 90.59 & 280.26 $\pm$ 12.89
        & 90.66 & 160.81 $\pm$ 4.42
        & 90.07 & 182.53 $\pm$ 7.41 \\
    CF-RNN~\cite{stankeviciute2021conformal}
        & 94.80 & 190.12 $\pm$ 3.13
        & 96.35 & 335.33 $\pm$ 16.73
        & 96.41 & 180.00 $\pm$ 3.97
        & 96.49 & 201.96 $\pm$ 5.39 \\
    Shuffle-Cov
        & 90.02 & 117.47 $\pm$ 4.41
        & 90.80 & 195.26 $\pm$ 27.27
        & 90.25 & 133.45 $\pm$ 8.25
        & 89.41 & 136.87 $\pm$ 13.57 \\
    \textbf{CRISP (Ours)}
        & 90.37 & \textbf{117.37 $\boldsymbol{\pm}$ 6.67}
        & 90.15 & \textbf{146.71 $\boldsymbol{\pm}$ 2.19}
        & 89.94 & \textbf{93.49 $\boldsymbol{\pm}$ 2.35}
        & 89.01 & \textbf{106.00 $\boldsymbol{\pm}$ 2.65} \\
\hline
    Action
    & \SetCell[c=2]{c,m} Polishing &
    & \SetCell[c=2]{c,m} Heavy P\&P &
    & \SetCell[c=2]{c,m} Light P\&P &
    & \SetCell[c=2]{c,m} Average & \\
\hline
    Method
    & Coverage & Efficiency $\downarrow$
    & Coverage & Efficiency $\downarrow$
    & Coverage & Efficiency $\downarrow$
    & Coverage & Efficiency $\downarrow$ \\
\hline
    MultiDimSPCI~\cite{xu2024conformal}
        & 90.42 & 106.54 $\pm$ 5.10
        & 89.51 & 133.33 $\pm$ 2.25
        & 90.51 & 123.06 $\pm$ 3.26
        & 90.00 & 122.52 $\pm$ 0.61 \\
    CopulaCPTS~\cite{sun2023copula}
        & 90.15 & 170.80 $\pm$ 4.73
        & 88.92 & 213.54 $\pm$ 6.59
        & 89.83 & 202.53 $\pm$ 6.06
        & 90.01 & 198.37 $\pm$ 3.36 \\
    CF-RNN~\cite{stankeviciute2021conformal}
        & 96.63 & 206.55 $\pm$ 12.99
        & 96.05 & 237.45 $\pm$ 2.20
        & 96.50 & 225.19 $\pm$ 5.14
        & 96.18 & 225.23 $\pm$ 3.31 \\
    Shuffle-Cov
        & 89.25 & 155.38 $\pm$ 44.72
        & 90.16 & 221.23 $\pm$ 23.61
        & 90.93 & 201.84 $\pm$ 22.75
        & 90.12 & 165.93 $\pm$ 10.03 \\
    \textbf{CRISP (Ours)}
        & 90.05 & \textbf{99.94 $\boldsymbol{\pm}$ 3.10}
        & 89.31 & \textbf{129.25 $\boldsymbol{\pm}$ 1.92}
        & 90.22 & \textbf{117.76 $\boldsymbol{\pm}$ 0.85}
        & 89.86 & \textbf{115.79 $\boldsymbol{\pm}$ 1.05} \\
\hline[0.8pt]
\end{tblr}
\vspace{-4mm}
\end{table*}

\begin{table}[t]
\centering
\vspace{-1mm}
\caption{Quantitative comparison on AMASS and 3DPW datasets.}
\label{tab:amass_3dpw_results}
\vspace{-2mm}
\footnotesize
\resizebox{\linewidth}{!}{%
\begin{tblr}
{
    colspec = {l *{2}{cc}},
    colsep  = 1pt,
    stretch = 0,
    rowsep  = 1.5pt,
    cells   = {halign=c, valign=m},
    column{1} = {halign=l, leftsep=5pt},
    vline{2,4} = {
        0.25pt,
        abovepos=-1,
        belowpos=-1
    },
    hborder{1-3,Z} = {
        abovespace=1.5pt,
        belowspace=1.5pt
    },
}
\hline[0.8pt]
    Dataset
    & \SetCell[c=2]{c,m} AMASS &
    & \SetCell[c=2]{c,m} 3DPW & \\
\hline
    Method
    & Coverage & Efficiency $\downarrow$
    & Coverage & Efficiency $\downarrow$ \\
\hline
    MultiDimSPCI~\cite{xu2024conformal}
        & 89.73 & 98.06 $\pm$ 1.69
        & 89.75 & 83.34 $\pm$ 0.87 \\
    CopulaCPTS~\cite{sun2023copula}
        & 89.79 & 444.64 $\pm$ 9.84
        & 90.07 & 366.32 $\pm$ 10.82 \\
    CF-RNN~\cite{stankeviciute2021conformal}
        & 98.11 & 650.28 $\pm$ 13.61
        & 98.57 & 560.46 $\pm$ 10.07 \\
    \textbf{CRISP (Ours)}
        & 89.79 & \textbf{95.93 $\boldsymbol{\pm}$ 2.08}
        & 90.03 & \textbf{82.73 $\boldsymbol{\pm}$ 0.93} \\
\hline[0.8pt]
\end{tblr}}
\vspace{-3mm}
\end{table}

\subsection{Motion Forecasting Benchmarks}
\noindent\textbf{Evaluation Metrics.}
We evaluate our method in terms of validity (\textit{Coverage}) and compactness (\textit{Efficiency}). \textit{Coverage} measures the percentage of test sequences where the ground-truth future poses are entirely enclosed by the prediction regions across all horizons, aiming to meet the target confidence level $1-\alpha$. \textit{Efficiency} calculates the average normalized volume of these prediction regions, where lower values indicate tighter and more precise uncertainty bounds.

\noindent\textbf{Results on Human3.6M.}
Table~\ref{tab:eqmotion_h36m_long} compares CRISP against representative conformal prediction baselines on Human3.6M. CRISP consistently yields the tightest prediction regions on 13 out of 15 actions while maintaining valid sequence-level coverage close to the target level. Specifically, CF-RNN produces overly conservative regions due to Bonferroni correction. 
Compared with MultiDimSPCI and CopulaCPTS, CRISP explicitly regularizes the uncertainty space with a skeletal prior to yield highly compact, geometrically aligned regions.
To further validate CRISP's model-agnostic nature, we evaluate all methods using PGBIG and siMLPe as base predictors, with and without TCL~\cite{tang2023temporal} training (Table~\ref{tab:alternative_predictors}). TCL progressively extends the prediction horizon through multi-stage training.
Across all predictor configurations, CRISP continues to achieve the best efficiency, demonstrating its robustness to different forecasting architectures and training strategies.
Notably, TCL slightly improves CRISP's efficiency with PGBIG, but not with siMLPe. This difference arises because the size of the prediction region also depends on the covariance geometry. 
Figure~\ref{fig:viz} visualizes the comparison results on Human3.6M. CRISP generates significantly tighter and more structurally aligned conformal ellipsoids that successfully enclose the ground truth, validating the efficacy of our topology-aware covariance in preventing spatial inflation. 
The visualization highlights interesting dynamics: temporally, the uncertainty volume progressively expands over longer horizons; spatially, highly constrained proximal root joints (e.g., hips) exhibit minimal variance compared to the high-degree-of-freedom distal extremities (e.g., wrists).

\noindent\textbf{Results on 3DPW and AMASS.}
Table~\ref{tab:amass_3dpw_results} summarizes the performance on AMASS and 3DPW.
CRISP robustly achieves valid empirical coverage and yields the most compact prediction regions on both datasets.
Notably, CRISP's efficiency gains over MultiDimSPCI are modest, potentially due to the extreme heterogeneity of motion patterns in the pooled calibration data, which may dilute the structural constraints provided by a shared, static topology prior. 

\noindent\textbf{Fairness.}
Unlike uncertainty-aware methods such as UnPOSed~\cite{saadatnejad2024toward} and SAGGB~\cite{wang2024existence}, which incorporate uncertainty modeling in training, CRISP calibrates prediction regions post hoc without retraining the predictor. Analyses of uncertainty types and calibration robustness are provided in the Additional Experiments section of the Appendix.

\begin{figure*}[t]
    \vspace{2mm}
    \centering
    \begin{minipage}[t]{0.49\textwidth}
        \centering
        \begin{minipage}[t]{0.49\linewidth}
            \centering
            \includegraphics[width=\linewidth]{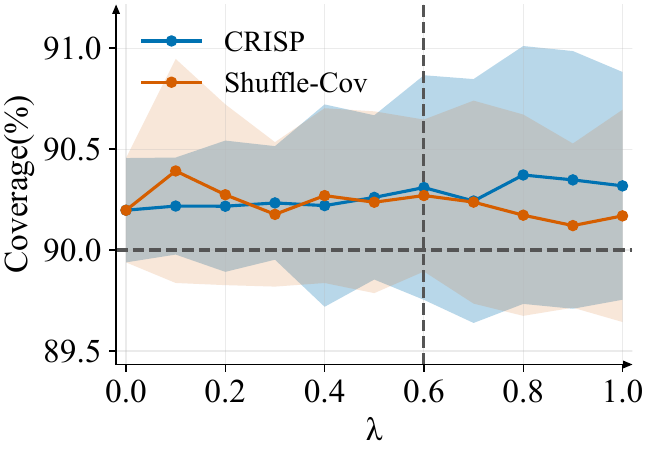}
        \end{minipage}
        \hfill
        \begin{minipage}[t]{0.49\linewidth}
            \centering
            \includegraphics[width=\linewidth]{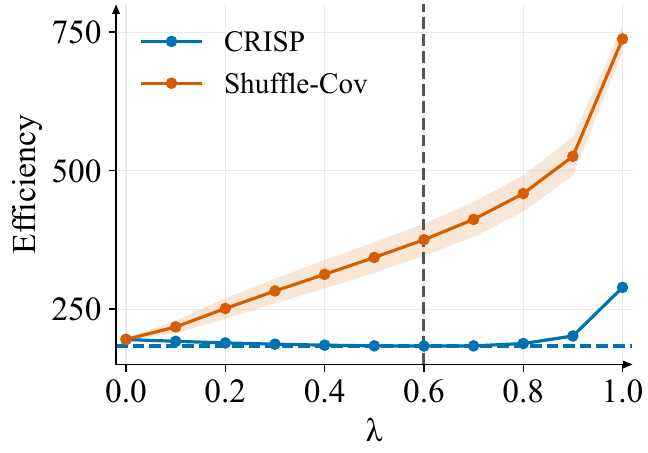}
        \end{minipage}
        \vspace{-1mm}
        {\small (a) Impact of $\lambda$ and skeletal prior on CRISP}
    \end{minipage}
    \hfill
    \begin{minipage}[t]{0.49\textwidth}
        \centering
        \begin{minipage}[t]{0.49\linewidth}
            \centering
            \includegraphics[width=\linewidth]{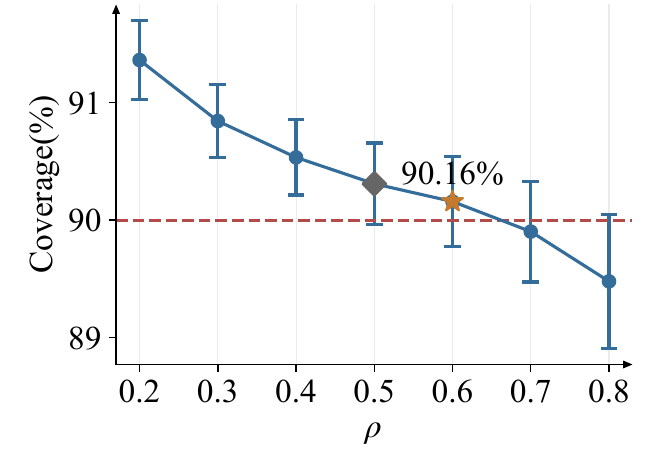}
        \end{minipage}
        \hfill
        \begin{minipage}[t]{0.49\linewidth}
            \centering
            \includegraphics[width=\linewidth]{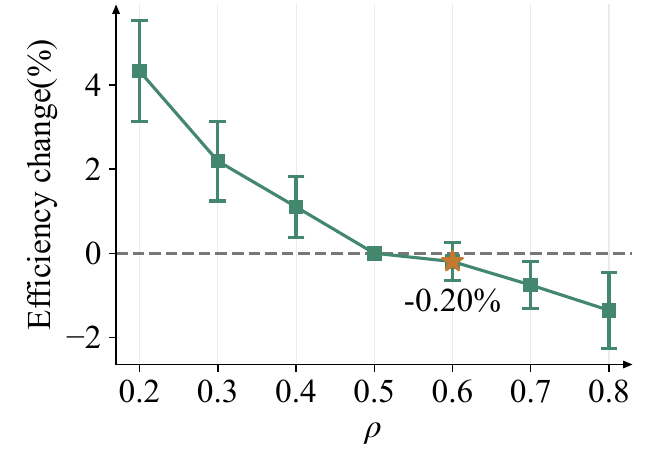}
        \end{minipage}
        \vspace{-1mm}
        {\small (b) Effect of calibration split}
    \end{minipage}
\vspace{-1mm}
\caption{(a) Impact of $\lambda$ and skeletal prior on coverage and efficiency. (b) Effect of calibration split ratio $\rho$. Left: empirical coverage (red dashed line denotes 90\% target); Right: relative efficiency change (negative values denote more compact regions).}
\vspace{-5mm}
\label{fig:ablation}
\end{figure*}

\subsection{Human-Robot Collaboration}
\noindent\textbf{Evaluation Metrics.}
To evaluate CRISP for collision warning in human-robot collaboration, we adapt the collision detection protocol from prior work. Specifically, a warning is triggered if the minimum distance between a human limb segment and a robotic arm link falls within a 20 cm threshold in any predicted frame, or if at least 20\% of either wrist's prediction region overlaps the danger zone. 

\noindent\textbf{Results on CHICO.}
Table~\ref{tab:chico_normal_short} presents the short-term prediction results on CHICO using SeS-GCN~\cite{sampieri2022pose} as the underlying predictor. Following~\cite{sampieri2022pose}, we evaluate motion forecasting on collision-free sequences. 
Across all actions, CRISP consistently yields smaller prediction regions than the baselines, with an average coverage close to the nominal level.

\begin{figure}[t]
\centering
\includegraphics[width=1.0\columnwidth]{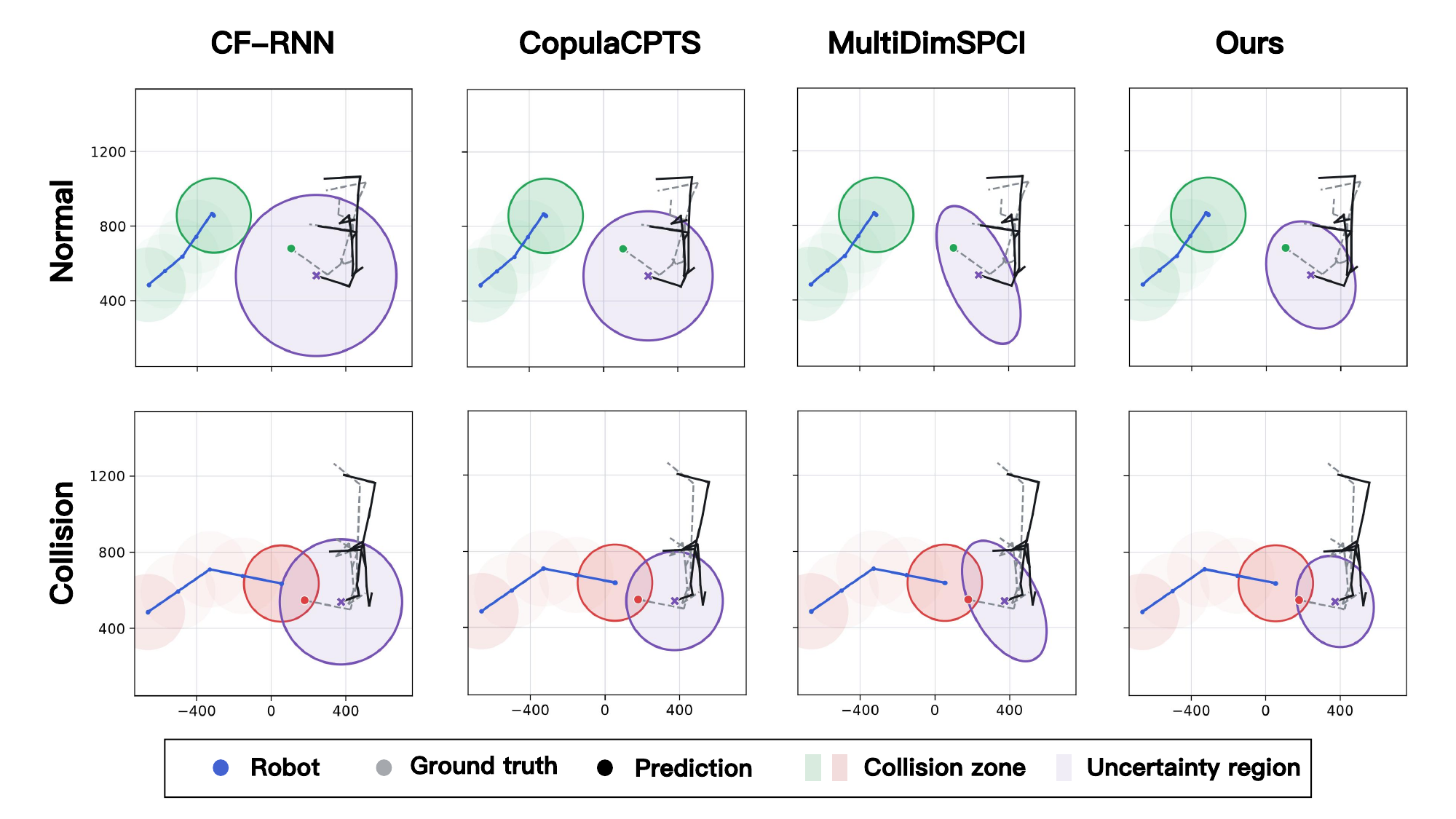} 
\vspace{-5mm}
\caption{Calibrated uncertainty regions at 400 ms for CHICO’s random pick and place task. Purple ellipses show the 3D regions projected onto the XY plane. Green (normal) and red (crash) mark the robot’s collision warning zone.}
\label{fig:chico_viz}
\end{figure}

\noindent\textbf{Collision Warning Evaluation.}
We evaluate collision warnings on 2 test subjects (S18 and S19) of CHICO. 
As shown in Table~\ref{tab:chico_collision}, CRISP improves recall from 0.47 to 0.68 and F1 score from 0.60 to 0.66, albeit with lower precision. By accounting for uncertainty bounds, CRISP can identify potential collision risks missed by point-based warnings. This trade-off is particularly valuable in safety-critical settings, where missed collision risks can have more serious consequences than unnecessary warnings.
Figure~\ref{fig:chico_viz} compares the 400\,ms wrist prediction regions for normal and crash sequences. In the collision case, the point forecast remains outside the warning zone even though the ground-truth wrist enters it. All four methods capture the collision risk, but CRISP achieves this with the most compact projected region, demonstrating that reliable risk identification does not require excessive spatial expansion.

\begin{table}[t]
    \centering
    \caption{Evaluation results of collision warnings on CHICO.}
    \label{tab:chico_collision}
    \vspace{-2mm}
    \small
    \begin{tblr}{
        colspec   = {lccc},
        colsep    = 6pt,
        stretch   = 0,
        rowsep    = 1.5pt,
        cells     = {valign=m},
        column{1} = {leftsep=5pt},
        vline{2}  = {0.25pt, abovepos=-1, belowpos=-1},
        hborder{1,2,Z} = {abovespace=1.5pt, belowspace=1.5pt},
        hline{1,Z} = {0.8pt},
        hline{2}   = {0.4pt},
    }
        Method          
        & Precision     
        & Recall        
        & F1 score \\
        SeS-GCN & \textbf{0.82} & 0.47 & 0.60 \\            
        SeS-GCN + CRISP & 0.65 & \textbf{0.68} & \textbf{0.66}
    \end{tblr}
    \vspace{-3mm}
\end{table}

\subsection {Ablation Study}
\noindent \textbf{Effect of Fusion Weight.} 
We evaluate CRISP's sensitivity to the fusion weight $\lambda$ between the empirical and topology-aware covariances (Figure~\ref{fig:ablation}(a)). 
The solid curves report the mean results and shaded regions indicate the corresponding variability across random seeds.
Coverage rate remains valid across $\lambda\in[0,1]$.
However, regional efficiency is optimized at $\lambda=0.6$, 
which demonstrates the complementarity of empirical residual statistics and skeletal topology priors in constructing compact prediction regions.

\noindent \textbf{Effect of Skeletal Topology.}
To assess the contribution of the skeletal prior, we compare the true skeletal distance matrix against Shuffle-Cov.
While both variants maintain target coverage (as shown in Figure~\ref{fig:ablation}(a)), the efficiency of the shuffled topology degrades severely as the weight of the prior ($\lambda$) increases. 
This confirms that our improvements stem from meaningful kinematic structures.

\noindent \textbf{Effect of Calibration Split.}
We investigate the effects of different calibration splits, denoted as $\rho=|\mathcal{D}_{\mathrm{cal}-1}|/(|\mathcal{D}_{\mathrm{cal}-1}|+|\mathcal{D}_{\mathrm{cal-2}}|)$.
As illustrated in Figure~\ref{fig:ablation}(b), increasing $\rho$ yields tighter prediction regions but at the expense of a gradual reduction in coverage. We identify $\rho=0.6$ (a $0.6/0.4$ split) as the optimal trade-off, maintaining valid coverage of $90.16\%$  while reducing the efficiency metric by $0.20\%$ compared to a balanced split.

\section{Conclusion}
We present CRISP, a framework for uncertainty quantification in human motion prediction. 
Under sequence-level exchangeability, CRISP provides distribution-free, finite-sample coverage guarantees. 
Its dual-calibration mechanism uses a skeletal prior to prevent spatial inflation and applies copula calibration to capture multi-step temporal dependencies.
Our theoretical analysis establishes coverage for the resulting prediction regions. Extensive experiments show that CRISP achieves coverage with remarkably tighter prediction regions compared to representative baselines.
{
    \small
    \bibliographystyle{ieeenat_fullname}
    \bibliography{main}
}


\end{document}